\documentclass{article}

\usepackage{microtype}
\usepackage{graphicx}
\usepackage{booktabs} 

\usepackage{hyperref}
\usepackage{tikz}
\usepackage{amsmath}
\usepackage{subfig}
\usepackage{makecell}
\usepackage{flafter} 

\usepackage[accepted]{icml2019}

\icmltitlerunning{Efficient data-driven encoding of scene motion using Eccentricity}

\begin{document}

\twocolumn[
\icmltitle{Efficient data-driven encoding of scene motion using Eccentricity}




\begin{icmlauthorlist}
\icmlauthor{Bruno Costa}{gfl}
\icmlauthor{Enrique Corona}{ric}
\icmlauthor{Mostafa Parchami}{ric}
\icmlauthor{Gint Puskorius}{ric}
\icmlauthor{Dimitar Filev}{ric}
\end{icmlauthorlist}

\icmlaffiliation{gfl}{Ford Greenfield Labs - Palo Alto, Ford Motor Company, Palo Alto, California, USA}
\icmlaffiliation{ric}{Research \& Innovation Center, Ford Motor Company, Dearborn, Michigan, USA}

\icmlcorrespondingauthor{Bruno Costa}{bcosta17@ford.com}
\icmlcorrespondingauthor{Enrique Corona}{ecoron18@ford.com}
\icmlcorrespondingauthor{Mostafa Parchami}{mparcham@ford.com}
\icmlcorrespondingauthor{Gint Puskorius}{gpuskori@ford.com}

\icmlkeywords{eccentricity analysis, eccentricity maps, time series, optical flow, dynamic scenes }

\vskip 0.3in
]



\printAffiliationsAndNotice{} 

\begin{abstract}
This paper presents a novel approach of representing dynamic visual scenes with static maps generated from video/image streams. Such representation allows easy visual assessment of motion in dynamic environments. These maps are 2D matrices calculated recursively, in a pixel-wise manner, that are based on the recently introduced concept of Eccentricity data analysis. Eccentricity works as a metric of discrepancy between a particular pixel of an image and its normality model, calculated in terms of mean and variance of past readings of the same spatial region of the image. While Eccentricity maps carry temporal information about the scene, actual images do not need to be stored nor processed in batches. Rather, all the calculations are done recursively, based on a small amount of statistical information stored in memory, thus resulting in a very computationally efficient (processor- and memory-wise) method. The list of potential applications include video-based activity recognition, intent recognition, object tracking, video description and so on.
\end{abstract}

\section{Introduction}
\label{sec:intro}

\begin{figure}[t]
\begin{centering}

\begin{tikzpicture}[picture format/.style={inner sep=3pt,}]

  \node[picture format]                   (A1)               {\includegraphics[width=0.7\columnwidth]{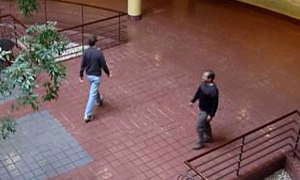}};
  \node[picture format,anchor=north]      (B1) at (A1.south) {\includegraphics[width=0.7\columnwidth]{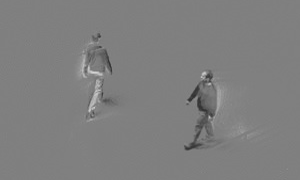}};

\end{tikzpicture}

\caption{The original frame of an example video followed by its resulting signed Eccentricity map with positive and negative components.}
\label{fig:eccexample} 
\end{centering}
\end{figure}

Time series (sequences of $n$-dimensional and equally spaced data observations) have been subject of study of many different areas. In contrast with other types of data structures, samples of a time series have a well-defined dependency relationship with other nearby samples. Analysis of time series are vastly used for prediction tasks, but also for classification and regression tasks where temporal information plays a crucial role.

For the specific domain of image streams, the challenge is to represent the dynamics and motion of objects in static images. In Figure \ref{fig:eccexample} \cite{MATLAB2017b} an example output of the technique proposed in this paper is presented. Such spatio-temporal descriptor images can be used as input to a variety of algorithms, allowing temporal information to be encoded into a model previously based on spatial data relationships.

Currently, there are a number of techniques that are able to capture patterns of apparent motion of objects in a visual scene caused by relative motion between an observer and a scene given a set of ordered images. Among the methods based on optical flow itself, one can mention the work of \cite{Wang2013}, where the authors introduce a static representation of videos based on dense trajectories and motion boundary descriptors. Such dense trajectories capture the local motion information of the video, enabling a good coverage of foreground motion. In \cite{bailer2015flow}, the authors present a variety of optical flow field implementations based on dense correspondence that is less outlier-prone and suitable for optical flow estimation, relying only on patch matching. In the specific application of action recognition, \cite{wu2011action} introduces an approach based on Lagrangian particle trajectories which, in turn, are a set of dense trajectories obtained by passing optical flow through time that enable representing the motion of a dynamic scene. The work of \cite{farnerback2003} presents a two-frame motion estimation algorithm that uses quadratic polynomials to approximate neighborhoods of each frame by applying a polynomial expansion transform. Then, they derive a method for correspondence fields by observing the translation of the polynomial transform. Finally, as an example of technique that is not based on optical flow, the authors of \cite{shi2015gradient} introduce the concept of gradient boundary histograms, an efficient local spatio-temporal descriptor built on simple spatio-temporal gradients, which are fast to compute.

However, most of these methods are very computationally expensive, which prevents their execution in real-time. Moreover, they may be highly sensitive to optimal choice of complex parameters, often requiring tuning by experts. Finally, they often heavily rely on feature extraction and matching between two consecutive frames, which often depends on consistent lighting conditions and absence of noise.

In this paper, we introduce a method for static representation of image series and apparent motion of objects, surfaces and edges that 

\begin{itemize}
\item Is efficient and able to process hundreds of frames per second on a standard computer;
\item is free of complex user-defined parameters and free of prior assumptions about the data and its distribution;
\item is inherently robust to process noise;
\item is able to handle concept drift, concept evolution;
\item can be used as input to solutions for problems that require temporal information (object tracking, human intent recognition, gesture recognition, video description and so on).
\end{itemize}

The algorithm proposed in this paper is based on the recently introduced concept of Data Eccentricity \cite{teda1}, which reflects how different a data point is from the past readings of the same set of variables. In other words, for a set of variables in an n-dimensional space, the value of the Eccentricity increases as the values for this set of variables deviate from their nominal behavior. Although it works very well for many different applications, the original concept of Eccentricity assumes an infinite memory (\emph{i.e.} it considers \emph{all} previous data samples when calculating the value for Eccentricity). In this proposal, we present an alternative formulation that is able to \emph{forget} older samples in order to create Eccentricity maps more suitable for dynamic scenes.

The remainder of this paper is structured as follows. In Section \ref{sec:teda}, we present the fundamentals of Eccentricity analysis. In Section \ref{sec:infinite}, we introduce our proposal for finite memory of Eccentricity and decomposition into positive and negative components. Then, the generation of Eccentricity maps from image streams is detailed in Section \ref{sec:eccmaps}. The experiment methodology and results are described in Section \ref{sec:results}. Finally, in Section \ref{sec:conclusion} the conclusions are presented.

\section{Eccentricity Data Analysis}
\label{sec:teda}

The original (infinite memory) concept of Eccentricity was introduced by \citet{teda1} and, since then, has been applied to many different tasks, such as clustering \cite{bezerra-eais2016}, classification \cite{angelov-tedaclass}, \cite{costa-wcci2016} and fault detection \cite{bezerra-eswa}. Eccentricity can be defined as a non-parametric, assumption-free methodology for extracting information from data \cite{Gu2017}. It is similar to traditional statistical learning and probability theory, however it does not assume random variables, nor pre-defined distributions, while being completely data-driven.

One crucial factor for the applicability of Eccentricity to a vast class of problems, including the ones in the field of computer vision, is the ability to recursively and incrementally update its knowledge basis. Such feature enables online learning, which is fundamental in dynamic environments, especially in the presence of data drift. Concept-drift and concept-evolution are often ignored by many of the state-of-the-art machine learning techniques. In the first case, an algorithm should be able to continuously adapt considering that the underlying concept of the data changes over time. In the second case, it should not assume that the structure of the data stream is fixed (\emph{e.g.} number of classes and number of clusters), since novel concepts may emerge when new data samples are available \cite{masud}.

The Eccentricity $\xi$ of the data sample $x$ at the time instant $k$ can be defined as~\cite{teda1}
\begin{equation}
\label{eq:teda1}
\xi_k=2\frac{\sum_{i=1}^{k}d(x,x_i)}{\sum_{i=1}^{k}\sum_{j=1}^{k}d(x_i,x_j)},
\end{equation}
\begin{equation}
k\geq 2, \qquad \sum_{i=1}^{k}\sum_{j=1}^{k}d(x_i,x_j)>0 \nonumber
\end{equation}
where $d$ is some type of distance (\emph{e.g.} Euclidean, Mahalanobis). Eccentricity is bounded~\cite{teda1} by
\begin{equation}
0\leq \xi_k \leq 1,\quad \sum_{i=1}^{k}\xi_k = 2\nonumber
\end{equation}

It should be highlighted that one of the main advantages of using Eccentricity as a similarity criterion is the ability update it recursively. Therefore, it has been shown that equation \ref{eq:teda1}, with $d$ being the Euclidean distance, can be derived as~\cite{teda1}
\begin{equation}
\label{eq:ecc}
\xi_k=\frac{1}{k}+\frac{(\mu_{k} - x_k)^{T}(\mu_{k} - x_k)}{k\sigma_k^2}
\end{equation}
\begin{equation}
\quad k\geq 2, \quad \sigma_k^2 > 0 \nonumber
\end{equation}
and both the mean $\mu_k$ and the variance $\sigma^2_k$ (unbiased form) can be recursively updated, respectively, by~\cite{teda1}
\begin{equation}
\label{eq:mean}
\mu_{k} = \frac{k-1}{k} \mu_{k - 1} + \frac{x_k}{k}
\end{equation}
\begin{equation}
\quad k\geq 1, \quad \mu_{1} = x_1 \nonumber
\end{equation}
\begin{equation}
\label{eq:sigma}
\sigma_{k}^2 = \frac{k-1}{k} \sigma^2_{k - 1} + \frac{(\mu_{k} - x_k)^{T}(\mu_{k} - x_k)}{k - 1}
\end{equation}
\begin{equation}
\quad \sigma^2_{1} = 0 \nonumber
\end{equation}

In practice, the recursive calculation of Eccentricity means that the data samples used to build the baseline model are not required to be stored in memory, since only very limited statistical information (mean and variance) are necessary the model update process. This brings substantial advantage to dynamic scenarios (where concept-drift and concept-evolution must be addressed) since it enables on-the-fly learning that can start from scratch (with the first data sample acquired) and theoretically allows an infinite number of samples to be aggregated in the model with no additional memory/computational cost.

For specific problems that require some type of hard classification (\emph{e.g.} detecting data samples that are significantly different from the learned baseline), a threshold is usually required. In the original proposal of Eccentricity calculation, the threshold to distinguish normal from anomalous data samples is based on the Chebyshev inequality~\cite{saw}, which states that, under any distribution, no more than $1/m^2$ of the data observations are more than $m\sigma$ away from the mean, where $\sigma$ represents the standard deviation of the data. Thus, a particular data sample $x_k$ is considered to be an anomaly if the condition
\begin{equation}
\xi_k > \frac{m^{2}+1}{2k}
\label{eq:threshold}
\end{equation}
\noindent is satisfied. The parameter $m$ is user-defined and directly affects the zone of influence of a data granule. Although it can be defined using multiple criteria, $m=3$ is largely used in literature \cite{liukkonen}, \cite{cook} as a standard value and presents satisfactory results for different data sets and different configurations.

\section{Finite Memory Eccentricity Calculation}
\label{sec:infinite}

In this paper we present a solution to a known issue of the original Eccentricity proposal: the infinite memory problem. Although sometimes desirable in specific frameworks, the fact that the Eccentricity calculation, as per equation \ref{eq:ecc}, takes into account all data samples back to $k = 1$ can be a significant problem in dynamic and rapidly evolving environments. In the original formulation, the influence of $x_k$ becomes decreasingly significant in the overall Eccentricity calculation as $k$ increases, up to the point where it is computationally null. This problem was partially addressed in \cite{bezerra2015rde} and is thoroughly addressed in this paper.

The idea is to introduce a finite memory mechanism for update of the Eccentricity that exponentially forgets older samples during the recursive calculation, making it more suitable for real-time processing of sequences of frames.

First, we replace the update factor $1/k$ by a constant forgetting factor $\alpha$ for the normality model (this parameter can be automatically extracted from data, but can safely be set to a small value, \emph{e.g.} $0.05$). The mean $\mu_k$ and variance $\sigma^2_k$ are now recursively updated respectively, by

\begin{equation}
\label{eq:meanfinite}
\mu_k = (1 - \alpha) \mu_{k-1} + \alpha x_k
\end{equation}
\begin{equation}
\label{eq:variancefinite}
\sigma^2_k = (1 - \alpha) \sigma^2_{k-1} + \frac{\alpha (x_k - \mu_k)^T (x_k - \mu_k)}{1 - \alpha}
\end{equation}

The introduction of a constant forgetting factor $0 \leq \alpha \leq 1$ assigns a set of exponentially decaying weights to the older observations $x_k$, as per
\begin{eqnarray}
W = [\alpha, \alpha(1 - \alpha)^{k - 1},\ ...\ , \alpha(1 - \alpha)^{k - K + 1}, ...] \nonumber
\end{eqnarray}
with unit sum. The vector $W$ forms a weighted average type aggregating operator with exponentially decaying weights that depend on $\alpha$.  The elements of $W$ with power greater than $K$ are approaching zero, hence defining a moving window of width $K$. It can be shown that the width of the moving window $K$ is approximately reciprocal to the forgetting factor $1/\alpha$.

The expression for Eccentricity was originally defined in equation \ref{eq:ecc} for of all data samples read up to the time instant $k$. By introducing the constant forgetting factor, the effect of the older data points (beyond $K$) is essentially eliminated.  Therefore, the Eccentricity can be expressed by the approximation: 

\begin{equation}
\label{eq:eccfinite}
\xi_k \approx \frac{1}{K}+\frac{(\mu_{k} - x_k)^{T}(\mu_{k} - x_k)}{K\sigma_k^2}
\end{equation}
and hence
\begin{equation}
\label{eq:eccfinite2}
\xi_k \approx \alpha + \frac{\alpha (x_k - \mu_k)^T (x_k - \mu_k)}{\sigma^2_k} 
\end{equation}

To prevent abrupt rise in $\xi_k$ when consecutive samples are very close to each other ($x_{k-n} \approx ... \approx x_{k-1} \approx x_k$, hence $\sigma^2_k \approx 0$), we introduce a variance threshold $\gamma$ within the Eccentricity calculation. The variance threshold is chosen such that the inherent noise in the sensor measurement does not show up as an eccentric signal. The Eccentricity formula then becomes
\begin{equation}
\label{eq:eccfinitegamma}
\xi_k = \alpha + \frac{\alpha (x_k - \mu_k)^T (x_k - \mu_k)}{max(\sigma^2_k, \gamma)} 
\end{equation}

For the anomaly detection case, the equation \ref{eq:threshold} can be rewritten in its finite memory form as
\begin{equation}
\xi_k > \frac{\alpha(m^{2}+1)}{2}
\label{eq:thresholdfinite}
\end{equation}

Finally, we can normalize $\xi_k$ so that its value always lie in the range $[0, 1[$. The normalized Eccentricity $\epsilon_k$ is then defined as
\begin{equation}
\label{eq:eccnormalized}
\epsilon_k = \frac{\xi_k- \alpha}{1 - \alpha} = \frac{\alpha (x_k - \mu_k)^T (x_k - \mu_k)}{(1 - \alpha)\ max(\sigma^2_k, \gamma)} 
\end{equation}

\subsection{Decomposition Into Components}

We can decompose Eccentricity as per definitions of equations \ref{eq:eccfinitegamma} and \ref{eq:eccnormalized} into positive and negative contributions. Such approach can be useful for understanding direction of the drift in values of the variables in the data stream.

Let $\xi^+_k$ and $\xi^-_k$ be the positive and negative components of the Eccentricity $\xi_k$. They are defined, respectively, by
\begin{equation}
\label{eq:eccpositive}
\xi^+_k =
\left \{
  \begin{tabular}{cc}
   $\alpha$, & $\left\Vert x_k \right\Vert^2 < \left\Vert \mu_k \right\Vert^2$ \\
   $\xi_k$, & $\left\Vert x_k \right\Vert^2 \geq \left\Vert \mu_k \right\Vert^2$
  \end{tabular}
\right.
\end{equation}
\begin{equation}
\label{eq:eccnegative}
\xi^-_k =
\left \{
  \begin{tabular}{cc}
   $\alpha$, & $\left\Vert x_k \right\Vert^2 \geq \left\Vert \mu_k \right\Vert^2$ \\
   $\xi_k$, & $\left\Vert x_k \right\Vert^2 < \left\Vert \mu_k \right\Vert^2$
  \end{tabular}
\right.
\end{equation}

Similarly, the normalized Eccentricity positive and negative components $\epsilon^+_k$ and $\epsilon^-_k$ are defined as
\begin{equation}
\label{eq:eccpositivenormalized}
\epsilon^+_k =
\left \{
  \begin{tabular}{cc}
   $0$, & $\left\Vert x_k \right\Vert^2 < \left\Vert \mu_k \right\Vert^2$ \\
   $\epsilon_k$, & $\left\Vert x_k \right\Vert^2 \geq \left\Vert \mu_k \right\Vert^2$
  \end{tabular}
\right.
\end{equation}
\begin{equation}
\label{eq:eccnegativenormalized}
\epsilon^-_k =
\left \{
  \begin{tabular}{cc}
   $0$, & $\left\Vert x_k \right\Vert^2 \geq \left\Vert \mu_k \right\Vert^2$ \\
   $\epsilon_k$, & $\left\Vert x_k \right\Vert^2 < \left\Vert \mu_k \right\Vert^2$
  \end{tabular}
\right.
\end{equation}

Finally, we can combine the negative and positive normalized components into a signed measure $\epsilon^*_k$ ranging between $[0, 1]$, defined as
\begin{equation}
\label{eq:eccsigned}
\epsilon^*_k = 0.5 + \frac{\epsilon^+_k - \epsilon^-_k}{2}
\end{equation}

\section{Eccentricity Maps from Image Streams}
\label{sec:eccmaps}

Eccentricity analysis have been proven very useful in many different fields of application. One of the fields that are still to be explored is image processing and computer vision. The finite memory formulation presented here makes the concept of Eccentricity suitable for processing of image streams, and due to its mathematical simplicity and very low required computational effort, it enables processing hundreds of images per second.

An Eccentricity map $E_k$ is a 2-D matrix of $a \times b$ dimensions, where $a$ and $b$ are the height and width of an original frame $I_k$, respectively. $E_k$ contains the spatio-temporal information regarding the changes in pixel intensities of the image stream $I$ over time, up to the time instant $k$, considering all image channels (in this case, let us consider a 3-channel RGB image).

Then, let each pixel $(i, j)$ of the image $I_k$ be the input $x_k$ for a separate Eccentricity unit. Let also ${R_k, G_k, B_k}$ be the three channels of the image $I_k$. Then we have
\begin{eqnarray}
x_k= \{R_k^{i,j},G_k^{i,j},B_k^{i,j}\} \nonumber
\end{eqnarray}
as the input vector containing the intensities of the three channels of a particular pixel $(i, j)$ of the image $I_k$ with $a \times b \times 3$ dimensions, at the given time instant $k$.

\begin{figure}[!ht]
\begin{centering}
\includegraphics[width=\columnwidth]{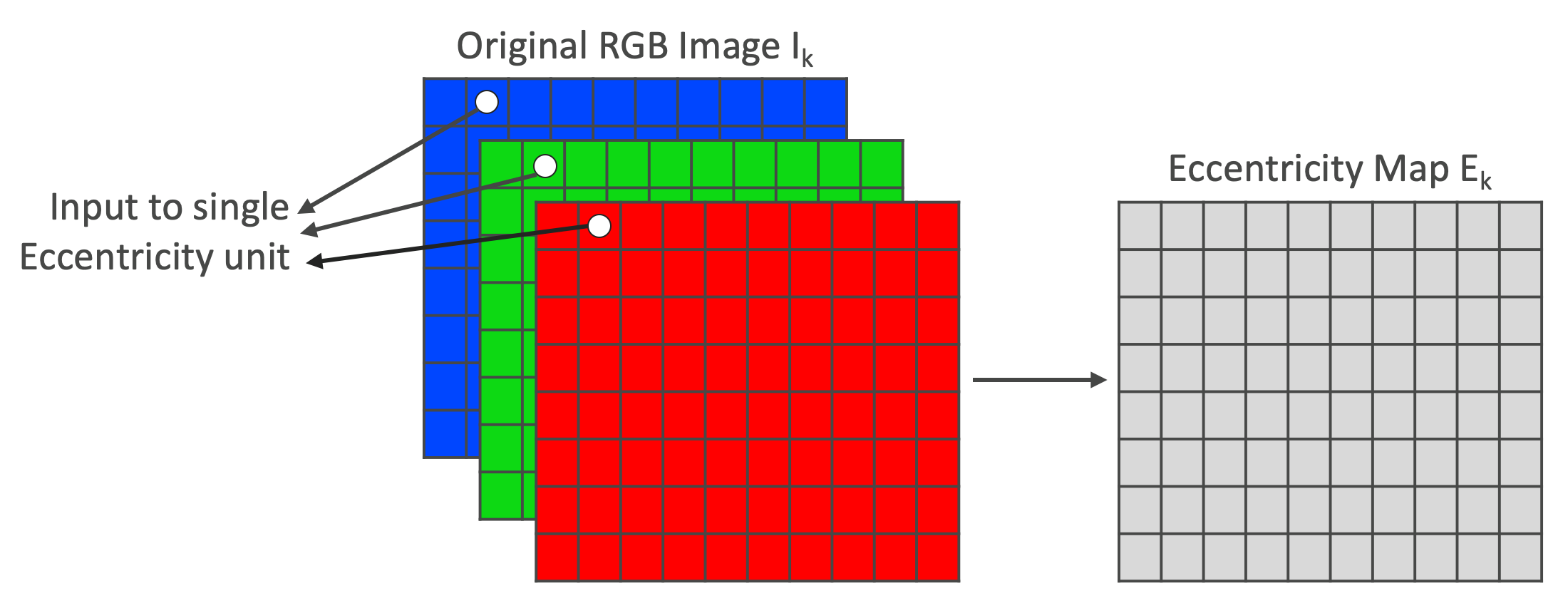}
\caption{Original image from image stream converted into Eccentricity map}
\label{fig:eccmap} 
\end{centering}
\end{figure}

In this formulation, each pixel is independent from the others and, hence, treated as a separate data stream. The Eccentricity map $E_k$, generated from the image $I_k$ with dimensions $a \times b \times 3$ (RGB), will have, in turn, $a \times b \times 1$ dimensions (gray scale), given that the Eccentricity calculation will result in a scalar $\epsilon_k$, regardless of the size of the input vector $x_k$, as per equation \ref{eq:eccnormalized}. An example of the Eccentricity map processing pipeline is given in Figure \ref{fig:eccmap}.

By translating equations \ref{eq:eccnormalized}, \ref{eq:eccpositivenormalized}, \ref{eq:eccnegativenormalized}, \ref{eq:eccsigned} and \ref{eq:thresholdfinite}, respectively, to the image domain, one can generate:

\begin{itemize}
\item {\bf Eccentricity maps ($E_k$): } aggregated measurement of change in pixel intensities through time, calculated in a pixel-wise manner.
\item {\bf Positive Eccentricity maps ($E^+_k$): } measurement of change in pixel intensities through time in the positive direction, \emph{i.e.} the new aggregated pixel intensity is higher than the moving mean for the same pixel.
\item {\bf Negative Eccentricity maps ($E^-_k$): } measurement of change in pixel intensities through time in the negative direction, \emph{i.e.} the new aggregated pixel intensity is lower than the moving mean for the same pixel.
\item {\bf Signed Eccentricity maps ($E^*_k$): } measurement of change in pixel intensities through time decomposed in positive and negative components.
\item {\bf Foreground mask ($F_k$): } Binary foreground/background segmentation mask based on the aggregated Eccentricity, Chebyshev threshold and closing operation.
\end{itemize}

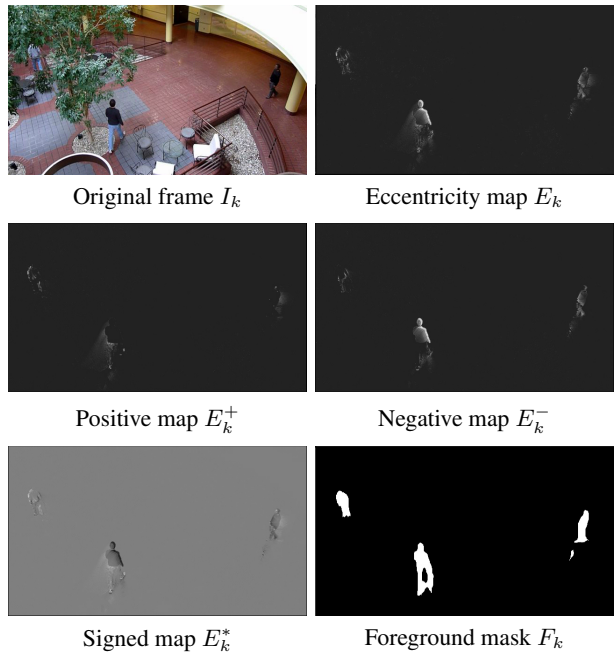
\begin{figure}
\centering
\begin{tikzpicture}[picture format/.style={inner sep=1.5pt,}]
  \node[picture format]                   (A1)               {\includegraphics[width=0.48\columnwidth]{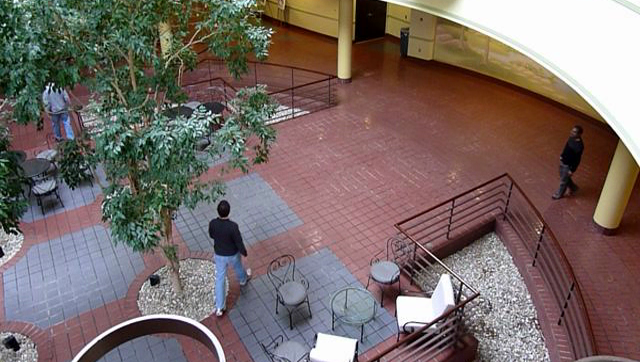}};
  \node[picture format,anchor=north west] (A2) at (A1.north east) {\includegraphics[width=0.48\columnwidth]{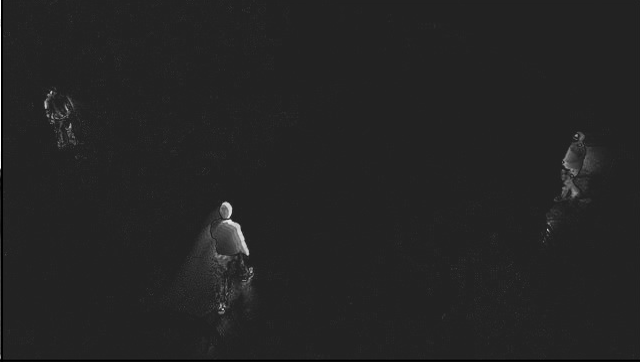}};
  \node[anchor=north] (B1) at (A1.south) {\footnotesize Original frame $I_k$};
  \node[anchor=north] (B2) at (A2.south) {\footnotesize Eccentricity map $E_k$};
  
  \node[picture format,anchor=north]      (C1) at (B1.south) {\includegraphics[width=0.48\columnwidth]{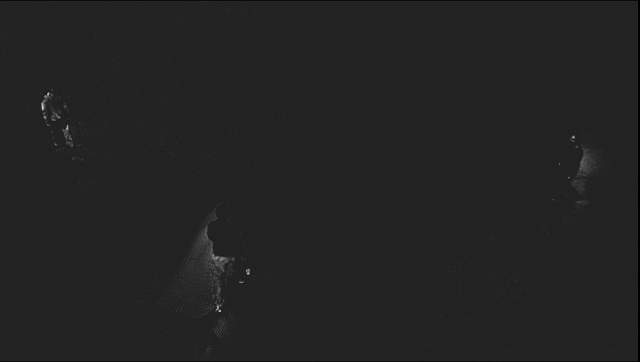}};
  \node[picture format,anchor=north]      (C2) at (B2.south)      {\includegraphics[width=0.48\columnwidth]{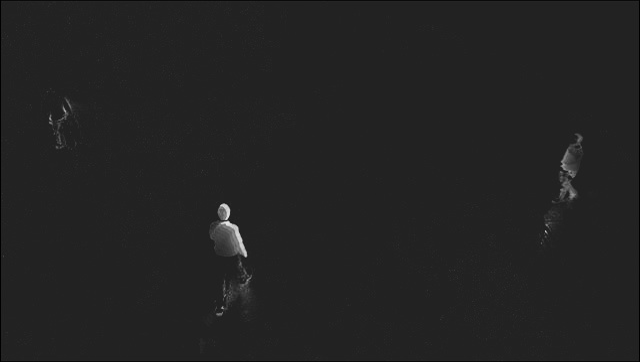}};
  \node[anchor=north] (D1) at (C1.south) {\footnotesize Positive map $E^+_k$};
  \node[anchor=north] (D2) at (C2.south) {\footnotesize Negative map $E^-_k$};

  \node[picture format,anchor=north]      (E1) at (D1.south) {\includegraphics[width=0.48\columnwidth]{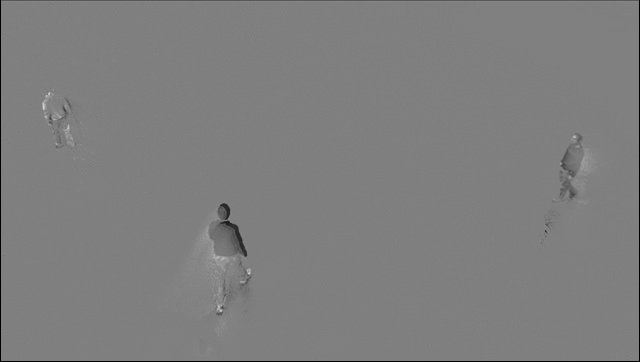}};
  \node[picture format,anchor=north]      (E2) at (D2.south)      {\includegraphics[width=0.48\columnwidth]{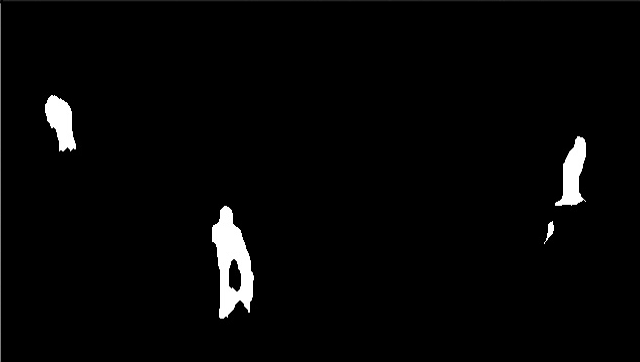}};
  \node[anchor=north] (F1) at (E1.south) {\footnotesize Signed map $E^*_k$};
  \node[anchor=north] (F2) at (E2.south) {\footnotesize Foreground mask $F_k$};
\end{tikzpicture}
\caption{Visual examples of all types of Eccentricity-based maps}
\label{fig:typesofmaps}
\end{figure}

Visual examples of all aforementioned types of maps are presented in Figure \ref{fig:typesofmaps} \cite{MATLAB2017b}.

\subsection{Background and Foreground Segmentation}

As illustrated in the bottom-right image in Figure \ref{fig:typesofmaps}, Eccentricity maps can be directly used for foreground/background image segmentation. Because moving objects directly change the intensities of the pixels affected by motion, the Eccentricity values naturally tend to increase in such areas of the image. By using a naturally dynamic threshold such as the Chebyshev inequality, an Eccentricity map can be classified into a foreground segmentation mask, which can be useful in many tasks such as fast and efficient detection of moving objects.

Although there are many existent techniques in literature that address this problem, such as \cite{gmm2} and \cite{bgs2}, which will be used later in this paper for validation, Eccentricity-based foreground/background segmentation has been shown \cite{costa2018evolving} to be a powerful tool due to its inherent ability to perform noise compensation ($\sigma^2$ and $\gamma$ in the denominator of equation \ref{eq:eccnormalized}) and its mathematical simplicity, which allows very fast calculations and enables real-time processing.

An Eccentricity-based foreground mask $F_k$ can be generated at the time instant $k$ as
\begin{equation}
\label{eq:foregroundmask}
F_k =
\left \{
  \begin{tabular}{cc}
   $0$, & $E_k \leq \displaystyle \frac{\alpha(m^{2}+1)}{2}$ \\ \\
   $1$, & $E_k > \displaystyle \frac{\alpha(m^{2}+1)}{2}$ \\
  \end{tabular}
\right.
\end{equation}
where $E_k$ is the 2D Eccentricity map at the time instant $k$ and $m$ is the user-defined parameter introduced in equation \ref{eq:threshold}. In the visual example presented in Figure \ref{fig:typesofmaps}, a \emph{closing} morphological operation is added to resulting foreground mask.

\section{Experiments and Results}
\label{sec:results}

In order to validate the proposal, we performed two sets of experiments. In the first experiment, we do a \emph{qualitative} comparison of signed Eccentricity maps and optical flow fields for activity recognition tasks. In the second experiment, we do a \emph{quantitative} assessment of Eccentricity maps in foreground/background segmentation tasks.

\subsection{Qualitative Comparison}

The technique proposed in this paper is used to generate signed Eccentricity maps that highlight moving objects in both positive and negative directions. The underlying concept of a signed Eccentricity map can be compared to, among other techniques, the one used in optical flow fields. The type of optical flow image we chose to use for comparison in this subsection was presented in \cite{wang2016temporal} and successfully used for activity recognition with temporal segment networks.

\begin{figure}[htp]
\centering
\subfloat[UCF101 - Juggling] {  
\begin{tikzpicture}[picture format/.style={inner sep=1.5pt,}]

  \node[picture format]                   (A1)               {\includegraphics[width=0.85in]{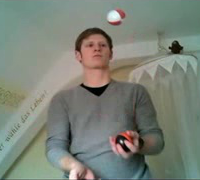}};
  \node[picture format,anchor=north]      (B1) at (A1.south) {\includegraphics[width=0.85in]{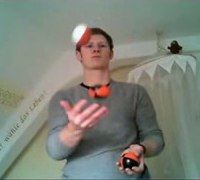}};
  \node[picture format,anchor=north]      (C1) at (B1.south) {\includegraphics[width=0.85in]{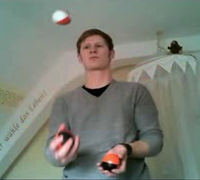}};

  \node[picture format,anchor=north west] (A2) at (A1.north east) {\includegraphics[width=0.85in]{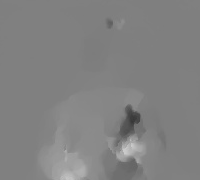}};
  \node[picture format,anchor=north]      (B2) at (A2.south)      {\includegraphics[width=0.85in]{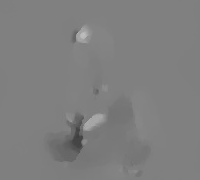}};
  \node[picture format,anchor=north]      (C2) at (B2.south)      {\includegraphics[width=0.85in]{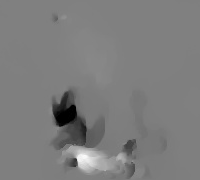}};

  \node[picture format,anchor=north west] (A3) at (A2.north east) {\includegraphics[width=0.85in]{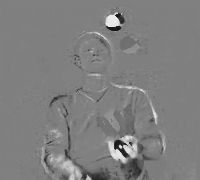}};
  \node[picture format,anchor=north]      (B3) at (A3.south)      {\includegraphics[width=0.85in]{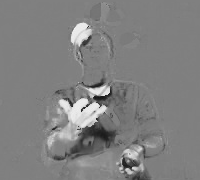}};
  \node[picture format,anchor=north]      (C3) at (B3.south)      {\includegraphics[width=0.85in]{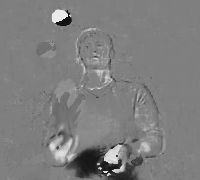}};

  \node[anchor=south] (D1) at (A1.north) {\footnotesize Original Frame};
  \node[anchor=south] (D2) at (A2.north) {\footnotesize Optical Flow};
  \node[anchor=south] (D3) at (A3.north) {\footnotesize Eccentricity Map};

\end{tikzpicture}
}
\\
\subfloat[UCF101 - Jumping Jack] {  
\begin{tikzpicture}[picture format/.style={inner sep=1.5pt,}]

   \node[picture format]                   (A1)               {\includegraphics[width=0.85in]{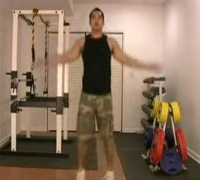}};
  \node[picture format,anchor=north]      (B1) at (A1.south) {\includegraphics[width=0.85in]{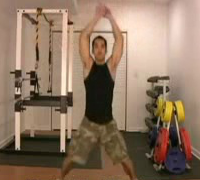}};
  \node[picture format,anchor=north]      (C1) at (B1.south) {\includegraphics[width=0.85in]{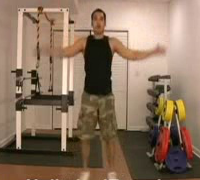}};

  \node[picture format,anchor=north west] (A2) at (A1.north east) {\includegraphics[width=0.85in]{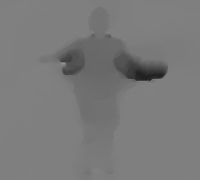}};
  \node[picture format,anchor=north]      (B2) at (A2.south)      {\includegraphics[width=0.85in]{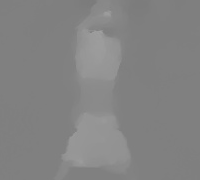}};
  \node[picture format,anchor=north]      (C2) at (B2.south)      {\includegraphics[width=0.85in]{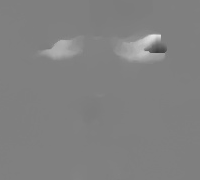}};

  \node[picture format,anchor=north west] (A3) at (A2.north east) {\includegraphics[width=0.85in]{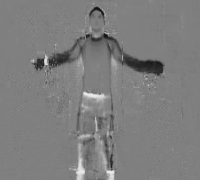}};
  \node[picture format,anchor=north]      (B3) at (A3.south)      {\includegraphics[width=0.85in]{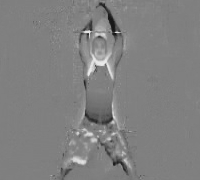}};
  \node[picture format,anchor=north]      (C3) at (B3.south)      {\includegraphics[width=0.85in]{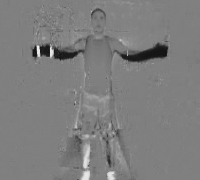}};

  \node[anchor=south] (D1) at (A1.north) {\footnotesize Original Frame};
  \node[anchor=south] (D2) at (A2.north) {\footnotesize Optical Flow};
  \node[anchor=south] (D3) at (A3.north) {\footnotesize Eccentricity Map};

\end{tikzpicture}
}
\\
\subfloat[UCF101 - Boxing] {  
\begin{tikzpicture}[picture format/.style={inner sep=1.5pt,}]

   \node[picture format]                   (A1)               {\includegraphics[width=0.85in]{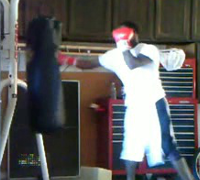}};
  \node[picture format,anchor=north]      (B1) at (A1.south) {\includegraphics[width=0.85in]{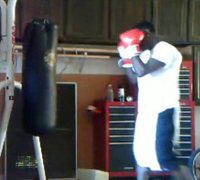}};
  \node[picture format,anchor=north]      (C1) at (B1.south) {\includegraphics[width=0.85in]{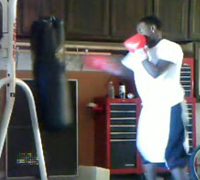}};

  \node[picture format,anchor=north west] (A2) at (A1.north east) {\includegraphics[width=0.85in]{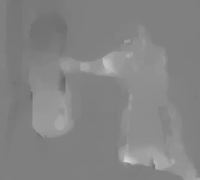}};
  \node[picture format,anchor=north]      (B2) at (A2.south)      {\includegraphics[width=0.85in]{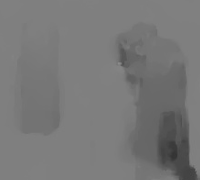}};
  \node[picture format,anchor=north]      (C2) at (B2.south)      {\includegraphics[width=0.85in]{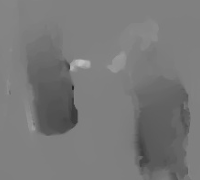}};

  \node[picture format,anchor=north west] (A3) at (A2.north east) {\includegraphics[width=0.85in]{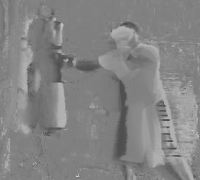}};
  \node[picture format,anchor=north]      (B3) at (A3.south)      {\includegraphics[width=0.85in]{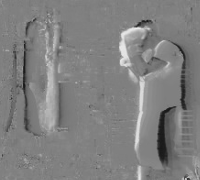}};
  \node[picture format,anchor=north]      (C3) at (B3.south)      {\includegraphics[width=0.85in]{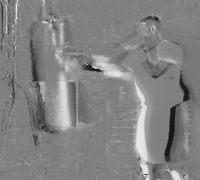}};

  \node[anchor=south] (D1) at (A1.north) {\footnotesize Original Frame};
  \node[anchor=south] (D2) at (A2.north) {\footnotesize Optical Flow};
  \node[anchor=south] (D3) at (A3.north) {\footnotesize Eccentricity Map};

\end{tikzpicture}
}
\caption{Visual comparison of optical flow images and signed Eccentricity maps on UCF101 dataset}
\label{fig:ucf101}
\end{figure}
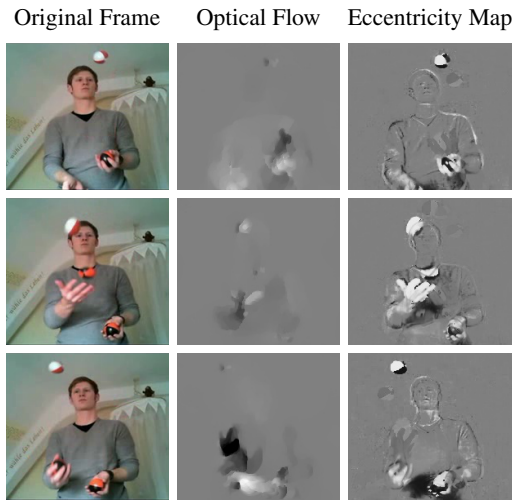
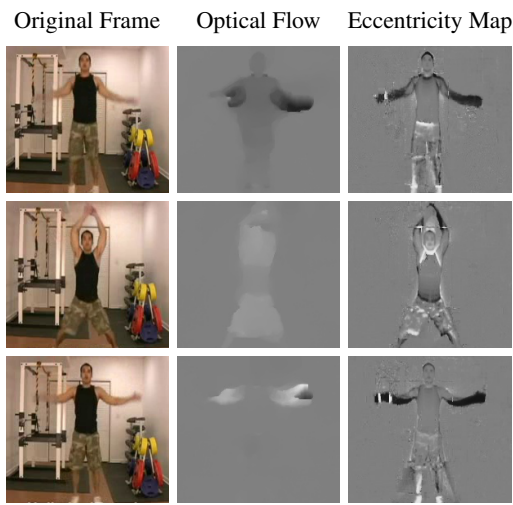
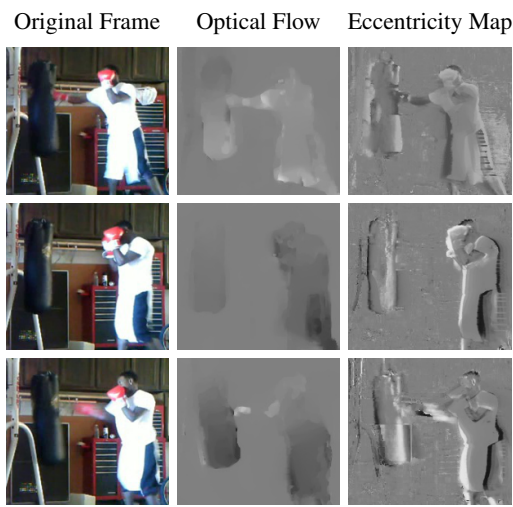

The dataset used for validation is the UCF101 and was introduced in \cite{soomro2012ucf101}. It consists of 101 action classes, thousands of clips and 27 hours of video data. The database is composed of realistic user uploaded videos containing camera motion and cluttered background. Some of the obtained results are presented in Figure \ref{fig:ucf101}.

By visual comparison, it is easy to note that, while optical flow images are indeed able to capture the basis of the motion through time, the Eccentricity map also incorporates some spatial information into its representation. In this manner, the images in the last column in each example aggregate spatio-temporal information in a single 1-channel image with significant levels of detail. For instance, in sequence (a), the motion trails of the hands and balls present in the Eccentricity map, together with the distinction between positive and negative components, give a clear notion of direction and magnitude of the motion. Similarly, in sequence (b), the negative component of Eccentricity, represented in the images by the darker pixels near the arms of the person, visually describes the vertical motion towards the ground.

\subsection{Quantitative Comparison}

In the second experiment, we use Eccentricity maps to create foreground segmentation masks that allow a straightforward comparison with similar techniques in literature.

For comparison purposes, we applied the proposed technique together with two well-known background subtraction algorithms (Eigen Subtraction \cite{bgs2} and Mixture of Gaussians \cite{gmm2}) to four datasets presented in \cite{i2r}. The implementation of the aforementioned two methods is provided by the BGSLibrary introduced in \cite{bgslibrary}.

For quantitative analysis, used the ground truth segmentation masks provided by the original authors and compared the resulting background/foreground mask of each algorithm with the ground truth mask, in a pixel-wise manner. Hence, an ideal segmentation would return a background/foreground mask that matches 100\% the ground truth mask in terms of true positives and true negatives. The metrics selected for evaluation are:

\begin{itemize}
\item TPR: True Positive Rate (recall)
\item PPV: Precision
\item ACC: Accuracy
\item F$_1$: F-score
\end{itemize}

\begin{figure}[t]
\centering
\subfloat[Dataset: Bootstrap] {  
\begin{tikzpicture}[picture format/.style={inner sep=1.5pt,}]

  \node[picture format]                   (A1)               {\includegraphics[width=0.9in]{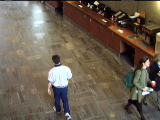}};
  \node[picture format,anchor=north]      (B1) at (A1.south) {\includegraphics[width=0.9in]{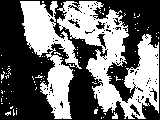}};

  \node[picture format,anchor=north west] (A2) at (A1.north east) {\includegraphics[width=0.9in]{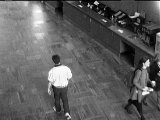}};
  \node[picture format,anchor=north]      (B2) at (A2.south)      {\includegraphics[width=0.9in]{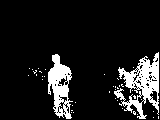}};

  \node[picture format,anchor=north west] (A3) at (A2.north east) {\includegraphics[width=0.9in]{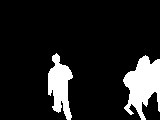}};
  \node[picture format,anchor=north]      (B3) at (A3.south)      {\includegraphics[width=0.9in]{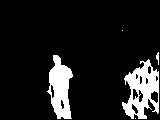}};

  \node[anchor=south] (D1) at (A1.north) {\footnotesize Original Frame};
  \node[anchor=south] (D2) at (A2.north) {\footnotesize Gray Scale};
  \node[anchor=south] (D3) at (A3.north) {\footnotesize Ground Truth};

  \node[anchor=north] (E1) at (B1.south) {\scriptsize Eigen Subtraction};
  \node[anchor=north] (E2) at (B2.south) {\scriptsize Mixture of Gaussians};
  \node[anchor=north] (E3) at (B3.south) {\scriptsize Eccentricity-based};
\end{tikzpicture}
}
\\
\subfloat[Dataset: Hall] {  
\begin{tikzpicture}[picture format/.style={inner sep=1.5pt,}]

  \node[picture format]                   (A1)               {\includegraphics[width=0.9in]{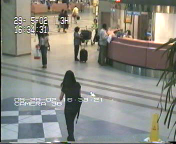}};
  \node[picture format,anchor=north]      (B1) at (A1.south) {\includegraphics[width=0.9in]{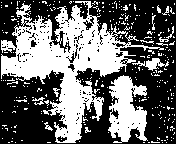}};

  \node[picture format,anchor=north west] (A2) at (A1.north east) {\includegraphics[width=0.9in]{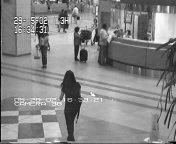}};
  \node[picture format,anchor=north]      (B2) at (A2.south)      {\includegraphics[width=0.9in]{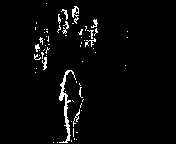}};

  \node[picture format,anchor=north west] (A3) at (A2.north east) {\includegraphics[width=0.9in]{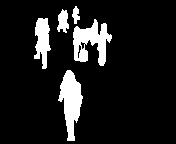}};
  \node[picture format,anchor=north]      (B3) at (A3.south)      {\includegraphics[width=0.9in]{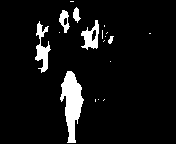}};

  \node[anchor=south] (D1) at (A1.north) {\footnotesize Original Frame};
  \node[anchor=south] (D2) at (A2.north) {\footnotesize Gray Scale};
  \node[anchor=south] (D3) at (A3.north) {\footnotesize Ground Truth};

  \node[anchor=north] (E1) at (B1.south) {\scriptsize Eigen Subtraction};
  \node[anchor=north] (E2) at (B2.south) {\scriptsize Mixture of Gaussians};
  \node[anchor=north] (E3) at (B3.south) {\scriptsize Eccentricity-based};
\end{tikzpicture}
}
\\
\subfloat[Dataset: Shopping Mall] {  
\begin{tikzpicture}[picture format/.style={inner sep=1.5pt,}]

  \node[picture format]                   (A1)               {\includegraphics[width=0.9in]{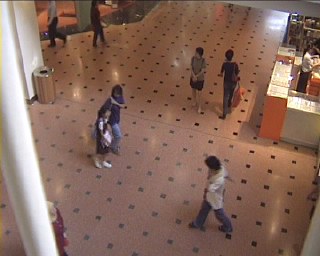}};
  \node[picture format,anchor=north]      (B1) at (A1.south) {\includegraphics[width=0.9in]{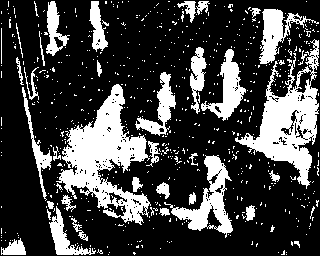}};

  \node[picture format,anchor=north west] (A2) at (A1.north east) {\includegraphics[width=0.9in]{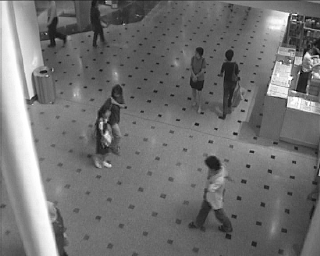}};
  \node[picture format,anchor=north]      (B2) at (A2.south)      {\includegraphics[width=0.9in]{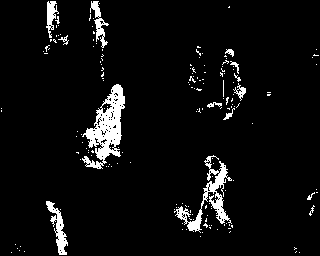}};

  \node[picture format,anchor=north west] (A3) at (A2.north east) {\includegraphics[width=0.9in]{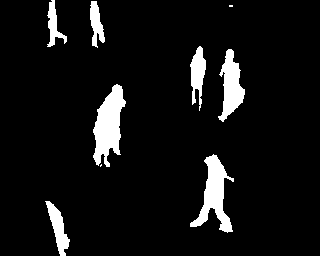}};
  \node[picture format,anchor=north]      (B3) at (A3.south)      {\includegraphics[width=0.9in]{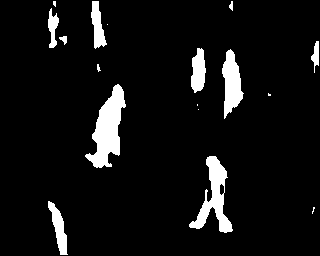}};

  \node[anchor=south] (D1) at (A1.north) {\footnotesize Original Frame};
  \node[anchor=south] (D2) at (A2.north) {\footnotesize Gray Scale};
  \node[anchor=south] (D3) at (A3.north) {\footnotesize Ground Truth};

  \node[anchor=north] (E1) at (B1.south) {\scriptsize Eigen Subtraction};
  \node[anchor=north] (E2) at (B2.south) {\scriptsize Mixture of Gaussians};
  \node[anchor=north] (E3) at (B3.south) {\scriptsize Eccentricity-based};
\end{tikzpicture}
}
\\
\subfloat[Dataset: Water Surface] {  
\begin{tikzpicture}[picture format/.style={inner sep=1.5pt,}]

  \node[picture format]                   (A1)               {\includegraphics[width=0.9in]{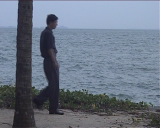}};
  \node[picture format,anchor=north]      (B1) at (A1.south) {\includegraphics[width=0.9in]{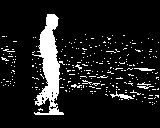}};

  \node[picture format,anchor=north west] (A2) at (A1.north east) {\includegraphics[width=0.9in]{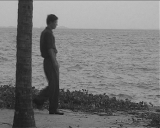}};
  \node[picture format,anchor=north]      (B2) at (A2.south)      {\includegraphics[width=0.9in]{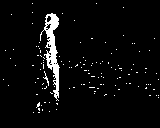}};

  \node[picture format,anchor=north west] (A3) at (A2.north east) {\includegraphics[width=0.9in]{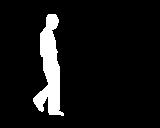}};
  \node[picture format,anchor=north]      (B3) at (A3.south)      {\includegraphics[width=0.9in]{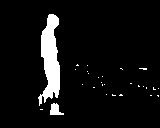}};

  \node[anchor=south] (D1) at (A1.north) {\footnotesize Original Frame};
  \node[anchor=south] (D2) at (A2.north) {\footnotesize Gray Scale};
  \node[anchor=south] (D3) at (A3.north) {\footnotesize Ground Truth};

  \node[anchor=north] (E1) at (B1.south) {\scriptsize Eigen Subtraction};
  \node[anchor=north] (E2) at (B2.south) {\scriptsize Mixture of Gaussians};
  \node[anchor=north] (E3) at (B3.south) {\scriptsize Eccentricity-based};
\end{tikzpicture}
}
\caption{Visual comparison of Optical Flow Fields and Eccentricity Maps on UCF101 Dataset}
\label{fig:resultssegmentation}
\end{figure}

The computer used to process the data is a 2.8 GHz Intel Core i7 with 16GB of RAM, and no GPU nor CUDA implementations were used in any of the experiments. The numerical results according to the forementioned metrics are presented in Table \ref{tab:results1}, while the comparison between execution times of each algorithm is presented in Table \ref{tab:results2}. Finally, examples of resulting background/foreground masks obtained with the three methods, including the one proposed in this paper, are presented in Figure \ref{fig:resultssegmentation}. 

\begin{table}[!th]
\centering
\caption{Quantitative Results - Accuracy Metrics}
\label{tab:results1} 
\begin{tabular}{ c l c l c l c l c }
\hline\noalign{\smallskip}
\multicolumn{5}{c}{\bf Dataset: Bootstrap (120$\times$160 px)} \\
\hline\noalign{\smallskip}
\makecell{{\bf Algorithm}} & \makecell{{\bf TPR}} & \makecell{{\bf PPV}} & \makecell{{\bf ACC}} & \makecell{{\bf F$_1$}} \\
\hline\noalign{\smallskip}
\makecell{Eigen\\ Subtraction} & \makecell{{\bf 0.88}} & \makecell{0.18} & \makecell{0.65} & \makecell{0.31} \\
\makecell{Mixture of\\ Gaussians}  & \makecell{0.41} & \makecell{0.70} & \makecell{0.93} & \makecell{0.52} \\
\makecell{Eccentricity\\ Maps}  & \makecell{0.53} & \makecell{{\bf 0.72}} & \makecell{{\bf 0.94}} & \makecell{{\bf 0.61}} \\ 
\hline\noalign{\smallskip}
\hline\noalign{\smallskip}
\multicolumn{5}{c}{\bf Dataset: Hall (176$\times$144 px)} \\
\hline\noalign{\smallskip}
\makecell{{\bf Algorithm}} & \makecell{{\bf TPR}} & \makecell{{\bf PPV}} & \makecell{{\bf ACC}} & \makecell{{\bf F$_1$}} \\
\hline\noalign{\smallskip}
\makecell{Eigen\\ Subtraction} & \makecell{{\bf 0.90}} & \makecell{0.16} & \makecell{0.70} & \makecell{0.28} \\ 
\makecell{Mixture of\\ Gaussians} & \makecell{0.32} & \makecell{{\bf 0.75}} & \makecell{0.95} & \makecell{0.45} \\
\makecell{Eccentricity\\ Maps} & \makecell{0.50} & \makecell{0.72} & \makecell{{\bf 0.96}} & \makecell{{\bf 0.59}} \\
\hline\noalign{\smallskip}
\hline\noalign{\smallskip}
\multicolumn{5}{c}{\bf Dataset: Shopping Mall (320$\times$256 px)} \\
\hline\noalign{\smallskip}
\makecell{{\bf Algorithm}} & \makecell{{\bf TPR}} & \makecell{{\bf PPV}} & \makecell{{\bf ACC}} & \makecell{{\bf F$_1$}} \\
\hline\noalign{\smallskip}
\makecell{Eigen\\ Subtraction} & \makecell{{\bf 0.84}} & \makecell{0.23} & \makecell{0.82} & \makecell{0.36} \\
\makecell{Mixture of\\ Gaussians} & \makecell{0.49} & \makecell{0.74} & \makecell{0.96} & \makecell{0.59} \\
\makecell{Eccentricity\\ Maps} & \makecell{0.74} & \makecell{{\bf 0.83}} & \makecell{{\bf 0.97}} & \makecell{{\bf 0.78}} \\
\hline\noalign{\smallskip}
\hline\noalign{\smallskip}
\multicolumn{5}{c}{\bf Dataset: Water Surface (160$\times$128 px)} \\
\hline\noalign{\smallskip}
\makecell{{\bf Algorithm}} & \makecell{{\bf TPR}} & \makecell{{\bf PPV}} & \makecell{{\bf ACC}} & \makecell{{\bf F$_1$}} \\
\hline\noalign{\smallskip}
\makecell{Eigen\\ Subtraction} & \makecell{{\bf 0.90}} & \makecell{0.58} & \makecell{0.94} & \makecell{0.71} \\
\makecell{Mixture of\\ Gaussians}  & \makecell{0.11} & \makecell{0.55} & \makecell{0.92} & \makecell{0.18} \\
\makecell{Eccentricity\\ Maps}  & \makecell{0.67} & \makecell{{\bf 0.90}} & \makecell{{\bf 0.97}} & \makecell{{\bf 0.77}} \\
\hline
\end{tabular}
\end{table}

\begin{table}[!th]
\centering
\caption{Quantitative Results - Execution Time (s)}
\label{tab:results2} 
\begin{tabular}{ c l c l c l c }
\hline\noalign{\smallskip}
\thead{\small\bf Dataset\\ \ } & \thead{\small\bf Bootstrap \\ \ } & \thead{\small\bf Hall \\ \ } & \thead{\small\bf Shopping\\ \small Mall} & \thead{\small\bf Water\\ \small Surface}\\
\hline\noalign{\smallskip}
\thead{\small\bf \# of \\ \small Frames} & \makecell{3055} & \makecell{3584} & \makecell{1286} & \makecell{633}\\
\hline\noalign{\smallskip}
\makecell{Eigen\\ Subtraction} & \makecell{4.312} & \makecell{6.253} & \makecell{6.800} & \makecell{0.970} \\
\makecell{Mixture of\\ Gaussians}  & \makecell{3.668} & \makecell{4.667} & \makecell{4.872} & \makecell{0.757} \\
\makecell{Eccentricity\\ Maps}  & \makecell{{\bf 0.778}} & \makecell{{\bf 1.168}} & \makecell{{\bf 1.754}} & \makecell{{\bf 0.142}} \\ 
\hline\noalign{\smallskip}

\hline
\end{tabular}
\end{table}

\section{Conclusion}
\label{sec:conclusion}

In this paper we proposed the application of Eccentricity analysis for generation of 2D Eccentricity maps to represent dynamic scenes in a static manner. Among the main advantages of Eccentricity maps, we discussed the detailed representation of spatio-temporal information in a 1-channel image and the low computational and memory requirements. The proposed technique was applied to different datasets for qualitative and quantitative comparison to other state-of-the-art techniques.

The visual comparison of signed Eccentricity maps with optical flow images revealed that former provides significantly more spatial information, while the temporal components clearly give a clear notion of direction and magnitude of motion. For the numerical analysis, we generated Eccentricity-based foreground segmentation masks and compared the results obtained with two other well-known techniques. The numerical results shown that the proposed technique provided significantly higher F-score in every case, in a small fraction of the computation time. Such numbers are very promising, especially when considering that the metrics were calculated in a pixel-wise manner.

As future work, we will use generated Eccentricity-maps as inputs to non-sequential neural networks for object detection, tracking, and intent prediction, and will develop a thorough study on how to quantify object motion based on Eccentricity maps.

\nocite{langley00}

\bibliography{bibliography}

\begin{thebibliography}{24}
\providecommand{\natexlab}[1]{#1}
\providecommand{\url}[1]{\texttt{#1}}
\expandafter\ifx\csname urlstyle\endcsname\relax
  \providecommand{\doi}[1]{doi: #1}\else
  \providecommand{\doi}{doi: \begingroup \urlstyle{rm}\Url}\fi

\bibitem[Angelov(2014)]{teda1}
Angelov, P.
\newblock Anomaly detection based on eccentricity analysis.
\newblock In \emph{Proc. {IEEE} Symposium Series in Computational Intelligence
  ({SSCI} 2014)}, Orlando, Florida, U.S.A., December 2014.

\bibitem[Bailer et~al.(2015)Bailer, Taetz, and Stricker]{bailer2015flow}
Bailer, C., Taetz, B., and Stricker, D.
\newblock Flow fields: Dense correspondence fields for highly accurate large
  displacement optical flow estimation.
\newblock In \emph{Proceedings of the IEEE international conference on computer
  vision}, pp.\  4015--4023, 2015.

\bibitem[Bezerra et~al.(2015)Bezerra, Costa, Guedes, and
  Angelov]{bezerra2015rde}
Bezerra, C.~G., Costa, B. S.~J., Guedes, L.~A., and Angelov, P.~P.
\newblock Rde with forgetting: An approximate solution for large values of $k$
  with an application to fault detection problems.
\newblock In \emph{International Symposium on Statistical Learning and Data
  Sciences}, pp.\  169--178. Springer, 2015.

\bibitem[Bezerra et~al.(2016{\natexlab{a}})Bezerra, Costa, Guedes, and
  Angelov]{bezerra-eais2016}
Bezerra, C.~G., Costa, B. S.~J., Guedes, L.~A., and Angelov, P.~P.
\newblock A new evolving clustering algorithm for online data streams.
\newblock In \emph{2016 IEEE Conference on Evolving and Adaptive Intelligent
  Systems (EAIS)}, pp.\  162--168, May 2016{\natexlab{a}}.

\bibitem[Bezerra et~al.(2016{\natexlab{b}})Bezerra, Costa, Guedes, and
  Angelov]{bezerra-eswa}
Bezerra, C.~G., Costa, B. S.~J., Guedes, L.~A., and Angelov, P.~P.
\newblock An evolving approach to unsupervised and real-time fault detection in
  industrial processes.
\newblock \emph{Expert Systems with Applications}, 63:\penalty0 134 -- 144,
  2016{\natexlab{b}}.
\newblock ISSN 0957-4174.

\bibitem[Cook et~al.(1997)Cook, Maxwell, Barnett, and Strauss]{cook}
Cook, G.~E., Maxwell, J.~E., Barnett, R.~J., and Strauss, A.~M.
\newblock Statistical process control application to weld process.
\newblock \emph{IEEE Transactions on Industry Applications}, 33\penalty0
  (2):\penalty0 454--463, Mar 1997.
\newblock ISSN 0093-9994.

\bibitem[Costa \& Goh(2018)Costa and Goh]{costa2018evolving}
Costa, B. S.~J. and Goh, M.
\newblock Evolving background subtraction for dynamic lighting scenarios.
\newblock In \emph{2018 IEEE International Conference on Fuzzy Systems
  (FUZZ-IEEE)}, pp.\  1--7. IEEE, 2018.

\bibitem[Costa et~al.(2016)Costa, Bezerra, Guedes, and Angelov]{costa-wcci2016}
Costa, B. S.~J., Bezerra, C.~G., Guedes, L.~A., and Angelov, P.~P.
\newblock Unsupervised classification of data streams based on typicality and
  eccentricity data analytics.
\newblock In \emph{2016 IEEE International Conference on Fuzzy Systems
  (FUZZ-IEEE)}, pp.\  58--63, July 2016.

\bibitem[Farneb{\"a}ck(2003)]{farnerback2003}
Farneb{\"a}ck, G.
\newblock Two-frame motion estimation based on polynomial expansion.
\newblock In Bigun, J. and Gustavsson, T. (eds.), \emph{Image Analysis}, pp.\
  363--370, Berlin, Heidelberg, 2003. Springer Berlin Heidelberg.
\newblock ISBN 978-3-540-45103-7.

\bibitem[Gu et~al.(2017)Gu, Angelov, Gutierrez, Iglesias, and Sanchis]{Gu2017}
Gu, X., Angelov, P.~P., Gutierrez, G., Iglesias, J.~A., and Sanchis, A.
\newblock Parallel computing teda for high frequency streaming data clustering.
\newblock In Angelov, P., Manolopoulos, Y., Iliadis, L., Roy, A., and Vellasco,
  M. (eds.), \emph{Advances in Big Data: Proceedings of the 2nd INNS Conference
  on Big Data, October 23-25, 2016, Thessaloniki, Greece}, pp.\  238--253.
  Springer International Publishing, 2017.

\bibitem[Kangin et~al.(2016)Kangin, Angelov, and Iglesias]{angelov-tedaclass}
Kangin, D., Angelov, P., and Iglesias, J.~A.
\newblock Autonomously evolving classifier tedaclass.
\newblock \emph{Information Sciences}, 366:\penalty0 1--11, 2016.

\bibitem[Li et~al.(2004)Li, Huang, Gu, and Tian]{i2r}
Li, L., Huang, W., Gu, I. Y.-H., and Tian, Q.
\newblock Statistical modeling of complex backgrounds for foreground object
  detection.
\newblock \emph{IEEE Transactions on Image Processing}, 13\penalty0
  (11):\penalty0 1459--1472, Nov 2004.
\newblock ISSN 1057-7149.

\bibitem[Liukkonen \& Tuominen(2004)Liukkonen and Tuominen]{liukkonen}
Liukkonen, T. and Tuominen, A.
\newblock A case study of spc in circuit board assembly: statistical mounting
  process control.
\newblock In \emph{2004 24th International Conference on Microelectronics (IEEE
  Cat. No.04TH8716)}, volume~2, pp.\  445--448 vol.2, May 2004.

\bibitem[Masud et~al.(2011)Masud, Gao, Khan, Han, and Thuraisingham]{masud}
Masud, M., Gao, J., Khan, L., Han, J., and Thuraisingham, B.
\newblock Classification and novel class detection in concept-drifting data
  streams under time constraints.
\newblock \emph{Knowledge and Data Engineering, IEEE Transactions on},
  23\penalty0 (6):\penalty0 859--874, June 2011.
\newblock ISSN 1041-4347.

\bibitem[Mathworks(2017)]{MATLAB2017b}
Mathworks.
\newblock \emph{{MATLAB R2017b and Computer Vision System Toolbox 8.0}}.
\newblock The Mathworks, Inc., Natick, Massachusetts, 2017.

\bibitem[Oliver et~al.(2000)Oliver, Rosario, and Pentland]{bgs2}
Oliver, N.~M., Rosario, B., and Pentland, A.~P.
\newblock A bayesian computer vision system for modeling human interactions.
\newblock \emph{IEEE Transactions on Pattern Analysis and Machine
  Intelligence}, 22\penalty0 (8):\penalty0 831--843, Aug 2000.
\newblock ISSN 0162-8828.

\bibitem[Saw et~al.(1984)Saw, Yang, and Mo]{saw}
Saw, J.~G., Yang, M., and Mo, T.~C.
\newblock Chebyshev inequality with estimated mean and variance.
\newblock \emph{The American Statistician}, 38\penalty0 (2):\penalty0 130--132,
  1984.

\bibitem[Shi et~al.(2015)Shi, Laganiere, and Petriu]{shi2015gradient}
Shi, F., Laganiere, R., and Petriu, E.
\newblock Gradient boundary histograms for action recognition.
\newblock In \emph{2015 IEEE Winter Conference on Applications of Computer
  Vision}, pp.\  1107--1114. IEEE, 2015.

\bibitem[Sobral(2013)]{bgslibrary}
Sobral, A.
\newblock {BGSLibrary}: An opencv c++ background subtraction library.
\newblock In \emph{IX Workshop de Visao Computacional (WVC'2013)}, Rio de
  Janeiro, Brazil, Jun 2013.
\newblock URL \url{https://github.com/andrewssobral/bgslibrary}.

\bibitem[Soomro et~al.(2012)Soomro, Zamir, and Shah]{soomro2012ucf101}
Soomro, K., Zamir, A.~R., and Shah, M.
\newblock Ucf101: A dataset of 101 human actions classes from videos in the
  wild.
\newblock \emph{arXiv preprint arXiv:1212.0402}, 2012.

\bibitem[Wang et~al.(2013)Wang, Kl{\"a}ser, Schmid, and Liu]{Wang2013}
Wang, H., Kl{\"a}ser, A., Schmid, C., and Liu, C.-L.
\newblock Dense trajectories and motion boundary descriptors for action
  recognition.
\newblock \emph{International Journal of Computer Vision}, 103\penalty0
  (1):\penalty0 60--79, May 2013.
\newblock ISSN 1573-1405.

\bibitem[Wang et~al.(2016)Wang, Xiong, Wang, Qiao, Lin, Tang, and
  Van~Gool]{wang2016temporal}
Wang, L., Xiong, Y., Wang, Z., Qiao, Y., Lin, D., Tang, X., and Van~Gool, L.
\newblock Temporal segment networks: Towards good practices for deep action
  recognition.
\newblock In \emph{European Conference on Computer Vision}, pp.\  20--36.
  Springer, 2016.

\bibitem[Wu et~al.(2011)Wu, Oreifej, and Shah]{wu2011action}
Wu, S., Oreifej, O., and Shah, M.
\newblock Action recognition in videos acquired by a moving camera using motion
  decomposition of lagrangian particle trajectories.
\newblock In \emph{Computer Vision (ICCV), 2011 IEEE International Conference
  on}, pp.\  1419--1426. IEEE, 2011.

\bibitem[Zivkovic \& Van Der~Heijden(2006)Zivkovic and Van Der~Heijden]{gmm2}
Zivkovic, Z. and Van Der~Heijden, F.
\newblock Efficient adaptive density estimation per image pixel for the task of
  background subtraction.
\newblock \emph{Pattern recognition letters}, 27\penalty0 (7):\penalty0
  773--780, 2006.

\end{thebibliography}
\bibliographystyle{icml2019}

\end{document}